\documentclass[journal]{IEEEtran}
\usepackage{cite}
\usepackage{amsmath,amssymb,amsfonts}
\usepackage{algorithmic}
\usepackage{graphicx}
\usepackage{textcomp}

\usepackage{lipsum}
\usepackage{multirow}
\usepackage{hhline}
\usepackage{caption}
\usepackage{subcaption}
\usepackage{hyperref}
\usepackage{lineno}
\usepackage[bottom]{footmisc}

\usepackage{xcolor}
\definecolor{RoyalBlue}{RGB}{65, 105, 225}

\newcommand{\ed}[1]{\textcolor{black}{#1}}

\def\BibTeX{{\rm B\kern-.05em{\sc i\kern-.025em b}\kern-.08em
    T\kern-.1667em\lower.7ex\hbox{E}\kern-.125emX}}

\begin{document}

\title{Contrastive Cross-Modal Pre-Training: A General Strategy for Small Sample Medical Imaging}

\author{Gongbo Liang, \IEEEmembership{Member, IEEE}, Connor Greenwell, \IEEEmembership{Student Member, IEEE}, \\
Yu Zhang, \IEEEmembership{Student Member, IEEE}, Xin Xing, \IEEEmembership{Student Member, IEEE}, Xiaoqin Wang, \\
Ramakanth Kavuluru,   and Nathan Jacobs, \IEEEmembership{Senior Member, IEEE}

\thanks{Manuscript initial submitted December 23, 2020; revised June 3, 2021, and August 24, 2021; accepted August 30, 2021.}

\thanks{This work was sponsored by Grant No. IIS-1553116 from the National Science Foundation, Grant No. IRG-19-140-31 from the American Cancer Society, and Grant No. P30CA177558 from the National Cancer Institute.}

\thanks{Gongbo Liang was with the Department of Computer Science, University of Kentucky, Lexington, KY, USA. Now, he is with the Department of Computer Science and Information Technology, Eastern Kentucky University, Richmond, KY, USA.
(e-mail: gongbo.liang@eku.edu).}

\thanks{Connor Greenwell, Yu Zhang, Xin Xing, and Nathan Jacobs are with the Department of Computer Science, University of Kentucky, Lexington, KY, USA (e-mail: \{connor.greenwell, y.zhang, xxi242, nathan.jacobs\}@uky.edu).}

\thanks{Xiaoqin Wang is with the Department of Radiology, University of Kentucky, Lexington, KY, USA (e-mail: xiaoqin.wang@uky.edu).}

\thanks{Ramakanth Kavuluru is with the Division of Biomedical Informatic, University of Kentucky, Lexington, KY, USA (e-mail: ramakanth.kavuluru@uky.edu).}
}

\markboth{Journal of \LaTeX\ Class Files,~Vol.~XX, No.~X, June~2021}%
{Shell \MakeLowercase{\textit{Liang et al.}}: Contrastive Cross-Modal Pre-Training}

\maketitle

\begin{abstract}
A key challenge in training neural networks for a given medical imaging task is the difficulty of obtaining a sufficient number of manually labeled examples. In contrast, textual imaging reports are often readily available in medical records and contain rich but unstructured interpretations written by experts as part of standard clinical practice. We propose using these textual reports as a form of weak supervision to improve the image interpretation performance of a neural network without requiring additional manually labeled examples. We use an image-text matching task to train a feature extractor and then fine-tune it in a transfer learning setting for a supervised task using a small labeled dataset.
The end result is a neural network that automatically interprets imagery without requiring textual reports during inference. 
We evaluate our method on three classification tasks and find consistent performance improvements, reducing the need for labeled data by 67\%--98\%.

\end{abstract}


\section{Introduction}
Convolutional neural networks (CNNs) have been promising tools for imaging analysis on various tasks~\cite{salem2016analyzing,zhai2018learning,zhang2019defense,su2020deep,liang2021alzheimer} and have been 
rapidly adopted in the medical imaging domain~\cite{gulshan2016development,liu2018multi,mihail2019automatic,zhang20192d,liang2020imporved,xing2020dynamic,ying2021multi}. However, robust CNNs are typically trained using large quantities of manually annotated data, such as images with discrete labels or pixel-level labels~\cite{he2016deep,huang2017densely,esteva2017dermatologist,liang2019ganai,hannun2019cardiologist,falk2019u,liang2020weakly}. Though large number of images are acquired each year in the medical domain, the number of images with manually annotated labels remains quite small due to the high annotation cost~\cite{litjens2017survey,yu2019clinical,willemink2020preparing}. For instance, the cancer imaging archive (TCIA), one of the largest collections of cancer images that is available for public download, hosted 127 datasets as of December 1$^{st}$, 2020; only six of them contain more than 1000 cases, and a majority (80 out of 127) have fewer than a hundred cases~\cite{tcia}. The limited number of manually annotated images presents a barrier to adopting modern deep learning techniques in the medical imaging domain~\cite{litjens2017survey,yu2019clinical,willemink2020preparing,wang2020inconsistent}.

\begin{figure*}
  \begin{subfigure}[b]{0.31\textwidth}
    ~~~~~~~~\includegraphics[width=.8\textwidth]{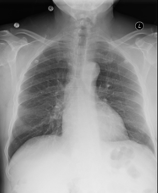}~~~~~~~~
  \end{subfigure}
  \begin{subfigure}[b]{0.63\textwidth}
    {\large \textbf{FINAL REPORT}}\\
    \tiny \smallskip
    \normalsize
    \textbf{HISTORY:} \_\_\_-year-old female with chest pain. \\
    \textbf{COMPARISON:} Comparison is made with chest radiographs from \_\_\_. \\
    \textbf{FINDINGS:} The lungs are well expanded. A retrocardiac opacity is seen which is likely due to atelectasis although infection is hard to exclude. Given the linear shape of the opacity, atelectasis is perhaps more likely. The heart is top-normal in size. The cardiomediastinal silhouette is otherwise unremarkable. There is no pneumothorax or pleural effusion.  Visualized osseous structures are unremarkable. \\
    \textbf{IMPRESSION:} Retrocardiac opacity, likely due to atelectasis but possibly due to pneumonia in the appropriate setting. \\
  \end{subfigure}
  \caption{An example of a chest x-ray (left) with the radiology report (right) from the MIMIC-CXR dataset.}
  \label{fig:data_example}  
\end{figure*}

In contrast to manually annotated labels, textual imaging reports, which are often readily available in medical records, contain interpretations written by experts as part of standard clinical practice (Figure~\ref{fig:data_example}). These unstructured textual data provide rich information about the images, but it is difficult to directly integrate the data into CNN training as labels. 

We propose using textual imaging reports as a weak supervision signal. Our model gleans the associations between text and images through a text and image matching network that we call TIMNet (Text-Image Matching Network).  
The network learns the image features from a large number of text-image pairs. These training pairs include both positive examples of images and associated reports and those that randomly pair an image with a report not associated with it. In this sense, our model learns to contrast cross-modal pairs that are valid matches with those that do not correspond to the same diagnostic imaging event. 
Various models for downstream application can then be built using the pre-trained image feature extractor.
Intuitively speaking, TIMNet transfers the strong expert generated language signal to the image feature part of the architecture, priming it more powerfully for downstream training on supervised tasks.
Since the feature extractor is pre-trained on a large and relevant dataset, transfer learning can be applied when building a downstream application that allows only a small labeled dataset to be used~\cite{mendel2019transfer,liang2019joint}.

Our method is widely applicable in the medical imaging domain, in which textual reports are often readily available but obtaining manual labels is expensive. We demonstrate our method on three classification tasks, but our method uses textual data in a general way that is not limited to specific image analysis tasks. Our experimental results show that the proposed method reduces the need for manually annotated data by up to 98\%.
Our contributions in this work are as follows: 
\begin{itemize}
    \item We propose a novel weakly supervised method of pre-training medical imaging analysis networks, which learns image features from a large but unlabeled dataset with associated textual reports;
    \item We demonstrate the proposed method via three classification tasks and also investigate the transferability of the learned features between datasets;
    \item We discuss the challenges of this study and provide a clear research direction for future researchers.
\end{itemize}

\section{Background}
\subsection{Textual Data in Image Analysis}

Researchers are actively seeking solutions for using textual data effectively in the field of medical imaging analysis. However, most works focus on deriving manual labels from textual data in different ways. For instance, CheXpert~\cite{irvin2019chexpert} and NegBio~\cite{peng2018negbio} propose two natural language processing (NLP) algorithms that are used for clinical text parsing. Both of them extract keywords (e.g., \texttt{pneumonia} and \texttt{pulmonary edema}) from the textual reports, then assign attributes (e.g., \texttt{positive} or \texttt{negative}) to the extraction keywords. The output of such algorithms are pairs of keywords and attributes, for instance, \texttt{positive pneumonia} or \texttt{negative pulmonary edema}. One may further infer discrete image-level labels from such attribute-keyword pairs and use the derived labels in CNN training~\cite{wu2020automatic,tam2020weakly,moradi2018bimodal,yuan2019automatic}.

Although such kinds of approaches are useful in automatically labeling a large amount of data, NLP models for label inference need to be trained separately, which increases the difficulty of using such methods. In addition, the accuracy of the derived labels could be an issue. As shown in~\cite{johnson2019mimic}, the automatically generated labels using CheXpert and NegBio do not always agree with each other in a large number of cases. Empirically speaking, the label inference errors   often manifest in a non-uniform manner, making the method even more problematic when training downstream application using the labels with nonuniform noise levels.

Instead of deriving labels from the clinical textual data, we propose using the textual data as weak supervision for image feature learning through text-image matching tasks. Our method uses textual data in a general way that is not limited to specific image analysis tasks. More importantly, there is no need to train a separate NLP model for label inferring that is independent of the image analysis tasks.

\subsection{Text and Image Matching} 

Text and image matching has become a popular research topic in recent years. One commonly used strategy is global representation matching, that usually involves three procedures: 1) image feature embedding, 2) text feature embedding, and 3) measures of the distance between the two embeddings. For instance, Kiros et al.~\cite{kiros2014unifying} used a CNN to encode images and a long short-term memory network (LSTM)~\cite{hochreiter1997long} to encode the full text. The triplet loss is used to pull the embeddings of the matched text and image closer to each other and push the unmatched ones further apart. Wehrmann et al.~\cite{wehrmann2018bidirectional} encoded the text data using an efficient character-level inception module that convolves over characters in the text. Sarafianos et al.~\cite{sarafianos2019adversarial} trained a text-image matching network by using a ResNet101~\cite{he2016deep} network for imaging processing and a transformer-based model~\cite{devlin2018bert} with an LSTM based encoder for text processing.

Most of the prior efforts focused on the matching problem from the perspective of text-based querying systems for image retrieval. In contrast, we aim to imbue image analysis models with the knowledge obtained from learning to match a medical image with the associated clinician-generated textual report.
We use the global matching approach with a two-branch setup one each for image processing and text processing. The absolute difference between the two output feature vectors is fed to a classification network, which predicts if the input image and text are a valid pair --- the text snippet is in fact the correct findings report for the image. 

\subsection{Contrastive Learning}
Contrastive learning is a machine learning strategy that learns the general features of a dataset by comparing the similarity and dissimilarity between data samples from the same class and different classes. In the imaging domain, contrastive learning can be done through Siamese networks that typically contain two identical subnetworks, which share the same weights~\cite{bromley1993signature,hadsell2006dimensionality,taigman2014deepface,schroff2015facenet}. The subnetworks take a pair of images either from the same class or different classes as input and output two feature vectors. A distance-based loss function---such as triplet loss~\cite{schultz2004learning,chechik2010large,schroff2015facenet} or contrastive loss~\cite{chen2020simple, he2020momentum, misra2020self}---is then used to compare the outputs that minimize the loss of samples from the same class and maximize the loss of samples from different classes.

Essentially, the Siamese network performs a binary classification task because each pair of input data is either from the ``same" class or ``not," which makes the binary cross-entropy loss a natural choice for Siamese network training. As triplet loss or contrastive loss, binary cross-entropy-based strategy has also been widely used~\cite{koch2015siamese,wu2018and,yang2020mscnn,shorfuzzaman2021metacovid,matlab}. For instance, Koch et al. firstly used such a method on image recognition and used binary cross-entropy loss to estimate the similarity between two feature vectors~\cite{koch2015siamese}; Wu et al. applied such a method on video-based person re-identification and got a promising result~\cite{wu2018and}; Yang et al. used binary cross-entropy to train a Siamese network that aims to overcome the extreme imbalance issue on text classification~\cite{yang2020mscnn}. In addition, binary cross-entropy-based training is also used by major scientific computation software companies; for instance MATLAB uses it in their contrastive learning tutorial~\cite{matlab}.

Contrastive learning is a natural choice to build our image and text matching network. We propose a Siamese-style network with the slightly variation where  the subnetworks have different architectures given the inherent variation in processing and representing text and image data. We use the binary cross-entropy method to form the distance-based learning model.

\subsection{Contextualized Natural Language Encoding}

Contextualized natural language embedding represents words as dense embeddings in a real vector space~\cite{mccann2017learned,peters2018deep,smith2020contextual}. Unlike the conventional word2vec embedding  method~\cite{mikolov2013distributed,mikolov2013efficient}, the embeddings of  contextualized methods are dependent on the surrounding context in which a word appears. The embeddings of the same word may be different according to the sentences in which the word occurs. Thus, contextualized embedding methods are more capable of addressing polysemy and homonymy issues in natural language processing. Popular contextualized embedding methods may include EMLo~\cite{peters2018deep}, transformers~\cite{vaswani2017attention}, and BERT~\cite{devlin2018bert}.

The transformer architecture~\cite{vaswani2017attention} is a well-known context-aware neural network architecture for natural language processing. It uses an encoder-decoder structure and attention mechanism to translate one sequence to another sequence. It encodes the contextual information of a word from distant parts of a sentence. In addition, unlike a conventional recurrent neural network (RNN), a transformer does not require sequential data to be processed in order, leading to new practical gains in efficiency. 

The \textit{bidirectional encoder representations from transformers} (BERT)~\cite{devlin2018bert} architecture trains a transformer by jointly conditioning on both left and right contexts in all layers. A pre-trained BERT model can be fine-tuned to create state-of-the-art models for various tasks without substantial task-specific architectural modifications~\cite{sun2019fine, lee2020biobert}. The BERT model contains twelve layers of transformer encoders. The outputs of these layers can be used as contextualized embeddings. In this study, we use a pre-trained BERT model as the contextualized natural language embedding model that encodes a given text as a feature vector. Specific details of this are presented in Section~\ref{sec-ws-train}.

\section{Method}
We assume two datasets ($X_P$ and $X_L$) exist, where $X_P$ has paired 2-tuples of images and associated textual reports and $X_L$ consists of labeled images. Specifically, $X_P =\{(x^{(j)}_i,x^{(j)}_t): x^{(j)}_i \in x_i, x^{(j)}_t \in x_t\}$ is an image and text paired dataset where $x_i$ is a set of images and $x_t$ is a set of textual notes such that $x^{(j)}_t$, the the $j$-report, corresponds to $x^{(j)}_i$, the $j$-th image. Also, $X_L=\{x_l, y\}$ is a labeled imaging dataset, where $x_l$ is a set of images and $y$ represents the set of corresponding labels, which could be image-level labels, bounding boxes, or pixel-level labels. Typically, we have  $|X_L| \ll |X_P| $.

We first learn an image encoding function $f$ using $X_P$ that encodes $x_i$ to a feature space. Then, $f$ can be used to build a downstream application using $X_L$ that maps $x_l$ to $y$. We learn $f$ through TIMNet, which predicts whether a given pair of $(x_i, x_t)$ is naturally related.

\begin{figure*}[!tb]
\centering
\includegraphics[width=0.925\textwidth]{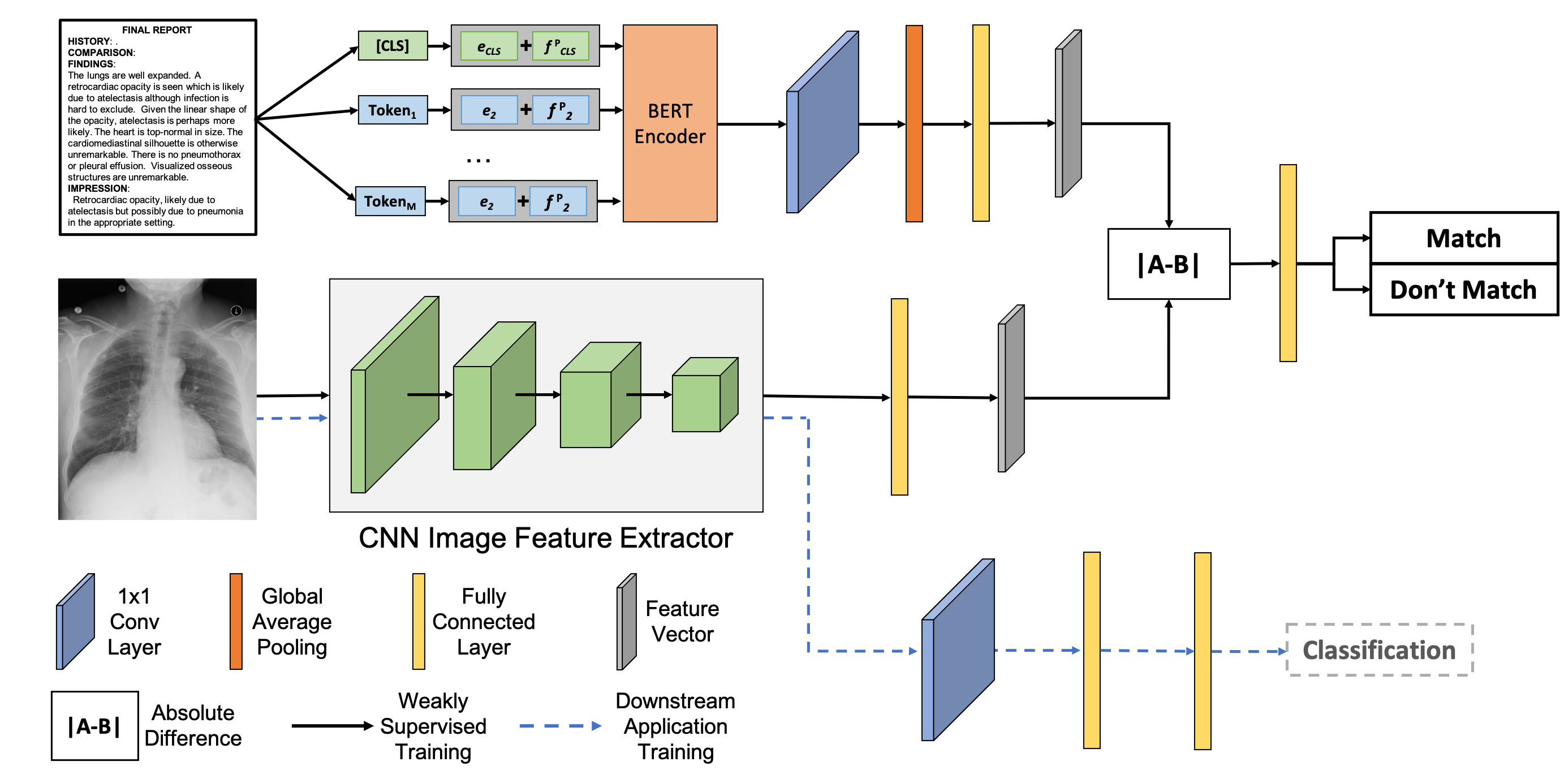}
\caption{The TIMNet cross-modal matching architecture. 1) weakly supervised image feature learning through a text-image matching network (solid black line). 2) Downstream application training using a small dataset (dashed blue line).}
\label{fig:architecture}
\end{figure*}

The proposed idea is illustrated in Figure~\ref{fig:architecture}. The network consists of two modules: 1) a weakly supervised image feature learning module and 2) a downstream application module for a specific task. The image encoding function $f$ is a CNN image feature extractor that is shared between the modules. 

\subsection{Weakly Supervised Image Feature Learning}
\label{sec-ws-train}
Weakly supervised image feature learning, i.e., image encoding is carried out through a global cross-modal matching approach with a two-branch network, one for text processing and the other for imaging processing. 
The input of the network is a text-image pair with a label indicating whether the text and image correspond to the same imaging event. The pair of the text and image is first projected into a vector space. Then, the absolute difference between the two output feature vectors is fed to a classification network, which predicts whether the input image and text are a true pair.

Mathematically, the text processing branch is defined as:
\begin{equation}
    v_t^{(j)} = h_t(x_t^{(j)}),
\end{equation}
where $h_t(\cdot)$ is the text processing network, $x_t^{(j)}$ is the $j^{th}$ input text, and $v_t^{(j)}$ is the output vector. The network is composed of a BERT encoder, a $1\times1$ convolutional (Conv) layer, a global average pooling (GAP) layer, and a fully-connected (FC) layer. The BERT encoder is used as a text feature extractor, which encodes the input text as contextualized embedding vectors. The vectors are then passed through the $1\times1$ Conv layer following GAP. 
The FC layer projects the vectors into a feature space as a feature vector. 
The image processing branch is defined as:
\begin{equation}
    v_i^{(j)} = h_{i}(x_i^{(j)}),
\end{equation}
where $h_i(\cdot)$ is the image processing network, $x_i^{(j)}$ is the $j^{th}$ input image, and $v_i^{(j)}$ is the output feature vector. The Conv layers of a ResNet model~\cite{he2016deep} are used as a feature extractor in $h_i(\cdot)$. An FC layer is added after the feature extractor that projects the image features into a feature space as a feature vector. 
The absolute difference of $v_t^{(j)}$ and $v_i^{(j)}$ is then passed to a shallow classification network, $h_{cls}(\cdot)$, for the matching prediction. Cross-entropy loss is used during the training of $h_{cls}(\cdot)$ to contrast valid matches from mismatched instances.

In general, the global matching network for weakly supervised image feature learning is expressed as:
\begin{equation}
    h(x^{(j)}) = h_{cls}(|h_t(x_t^{(j)}) - h_i(x_i^{(j)})|),
\end{equation}
where $x^{(j)}$ is the $j^{th}$ input, $x^{(j)}=\{x_t^{(j)}, x_i^{(j)}\}$, $x^{(j)} \in X_P$, $|\cdot|$ denotes the absolute difference, and $h(\cdot)$ is the whole matching network.
The absolute difference combining with cross-entropy loss forms a contrastive-style network that learns image features under a weakly supervised fasion. 

\subsection{Fine-Tuning for Downstream Tasks}
At the end of the weakly supervised feature learning phase described earlier, the hypothesis is that the matching process has pre-trained the parameterized image feature extractor $f$ to the extent that it needs fewer supervised instances for a downstream image-related task. This, in turn, relies on our high-level intuition that there is a nontrivial transferable signal available in the textual annotations to improve imaging tasks down the line.

The fine-tuning of the image processing branch in TIMNet for downstream tasks is fairly straightforward. Although it can be done for a variety of applications, in this study, we demonstrate the proposed method for classification tasks.

To build a downstream application model, we can either add additional Conv layers and FC layers to the image processing branch $h_i(\cdot)$, or use it as-is by retraining the FC layers. Since we need to optimize only the few additional layers from scratch, while the rest of the network is already pre-trained on a larger set of images from the same domain, the total number of required training instances for fine-tuning can be much smaller than training the entire network from scratch. 

\subsection{Implementation}
\label{sec:implementation}
The HuggingFace BERT with \texttt{bert-base-uncased} weights~\cite{wolf2019transformers} is used as the backbone of the text processing branch. A $1\times1$ Conv layer, a GAP layer, and an FC layer with 512 neurons are added to the BERT encoder. The ResNet-18 model with ImageNet pre-trained weights is used as the backbone of the imaging processing branch. A $1\times 1$ Conv layer and an FC layer with 512 neurons are added before and after the GAP layer, respectively. Both of the $1\times1$ Conv layers in the text processing and image processing branches are followed by batch normalization~\cite{ioffe2015batch} and a rectified linear unit (ReLU)~\cite{nair2010rectified}.

All the Conv layers of the image processing branch are used as a feature extractor in the downstream networks. A shallow CNN classification network is added on top of the feature extractor. The CNN classification network contains a $1\times 1$ Conv layer, an FC layer with 512 neurons, and an output layer with various numbers of neurons for different tasks. Cross-entropy loss and an Adam optimizer~\cite{kingma2014adam} with a learning rate of $10^{-4}$ are used for both the weakly supervised feature learning stage and the downstream application training stage.

\section{Evaluation and Discussion}

\subsection{Datasets}
We use the MIMIC-CXR~\cite{johnson2019mimic} and Mendeley-V2~\cite{kermany2018labeled} datasets in this study. The MIMIC-CXR dataset is used in both the TIMNet pre-training and downstream application training. Mendeley-V2 is used only for downstream application training. 

\subsubsection{MIMIC-CXR} 
This dataset contains 227,835 radiographic studies of 64,588 patients with 368,948 chest X-rays and the associated radiology reports. The dataset also provides 14 labels (with 13 labels for different abnormalities and one label for the normal case), which are derived from the radiology reports using NLP tools, such as NegBio and CheXpert~\cite{peng2018negbio,irvin2019chexpert}. In the official training/validation/testing split, the validation set and test set have 2,991 and 5,159 images, respectively. We combine the official validation and test sets to form our validation set. The chest X-ray images are resized to $500\times 500$. We use the official training set and our validation set in both TIMNet pre-training and downstream application training.

\subsubsection{Mendeley-V2} 
This dataset is a pediatric chest X-ray dataset that includes 4,273 pneumonia images and 1,583 normal images. Although the imaging modality is the same as that of MIMIC-CXR, the patient demographics are different. In addition, the images in Mendeley-V2 may have been acquired with devices from different vendors. We used the original training/validation split of this dataset to train a downstream application model that evaluates the transferability of pre-trained weights between different datasets while retaining the same imaging modality.

\subsection{Evaluation Setup}
We evaluate the proposed method through the downstream application performance and the degree of need for labeled instances for supervision in  downstream application training

\subsubsection{TIMNet Pre-Training}
The TIMNet is pre-trained for 50 epochs using both the true text-image pairs and negative pairs from the training set of MIMIC-CXR. A true pair contains an image and the associated report. A negative pair contains an image and a report randomly selected from the training set. The “findings” portion of the radiology reports is used as the text input. The length of each piece of text is preprocessed to 256 words at the word embedding stage. We add 0s for texts shorter than 256 words, and we snip much longer texts to 256 words. The output of this first weak-supervision phase is a probability estimate of the input text-image pair being a true match. All the Conv layers of the image processing branch are used as a feature extractor in the downstream networks. Shallow CNN classification networks are added on top of the feature extractor with the same architecture discussed in Sec~\ref{sec:implementation} for various downstream applications.

\subsubsection{Downstream Applications}
The downstream networks are trained using varying amounts of the training data and image-level labels from the MIMIC-CXR and Mendeley-V2 datasets, ranging from $0.5\%$ to $100\%$. Three downstream applications are trained, namely, 1) a binary classification model for abnormality diagnosis of MIMIC-CXR; 2) a multi-label classification model for lung disease diagnosis of MIMIC-CXR; and 3) a binary classification model for pediatric pneumonia diagnosis of Mendeley-V2. The downstream applications aim to test the TIMNet pre-trained feature extractor on the pre-training dataset (i.e., MIMIC-CXR) and an external dataset (i.e., Mendeley-V2). Each downstream network is trained for 100 epochs with a batch size of 16. No text is needed for training the downstream networks. 
We use the accuracy (ACC), the area under the receiver operating characteristic curve (auROC), precision (Prec), recall (Recall), F1 score (F1), and average precision (AP) as the evaluation metrics for the binary classification tasks and the auROC and AP for the multi-label classification tasks.

\subsection{Classification on Pre-Training Dataset}

We first present the evaluation results of downstream applications that are trained using the MIMIC-CXR dataset by comparing the proposed \ed{pre-training method (\textit{Ours}) with two other methods, \textit{Base} and \textit{ImageNet}. The \textit{Ours} model is pre-trained on MIMIC-CXR using the proposed text-image matching method. The \textit{Base} model is randomly initialized using the
default settings. The \textit{ImageNet} model is pre-trained on the ImageNet dataset~\cite{deng2009imagenet}. We use the PyTorch provided ImageNet weights in this study.} 

Two downstream applications are trained and tested \ed{using the MIMIC-CXR dataset}: a binary classification model and a multi-label classification model. The binary classification model predicts whether an abnormality exists in an image, and the multi-label classification model predicts what kind of abnormality exists in an image. The output of the binary classification model is Boolean, and the output for the multi-label task is a multi-hot vector, with 1 indicating the presence of a particular class.

\subsubsection{Binary Classification}
\begin{figure}[!tb]
\centering
\includegraphics[width=0.5\textwidth]{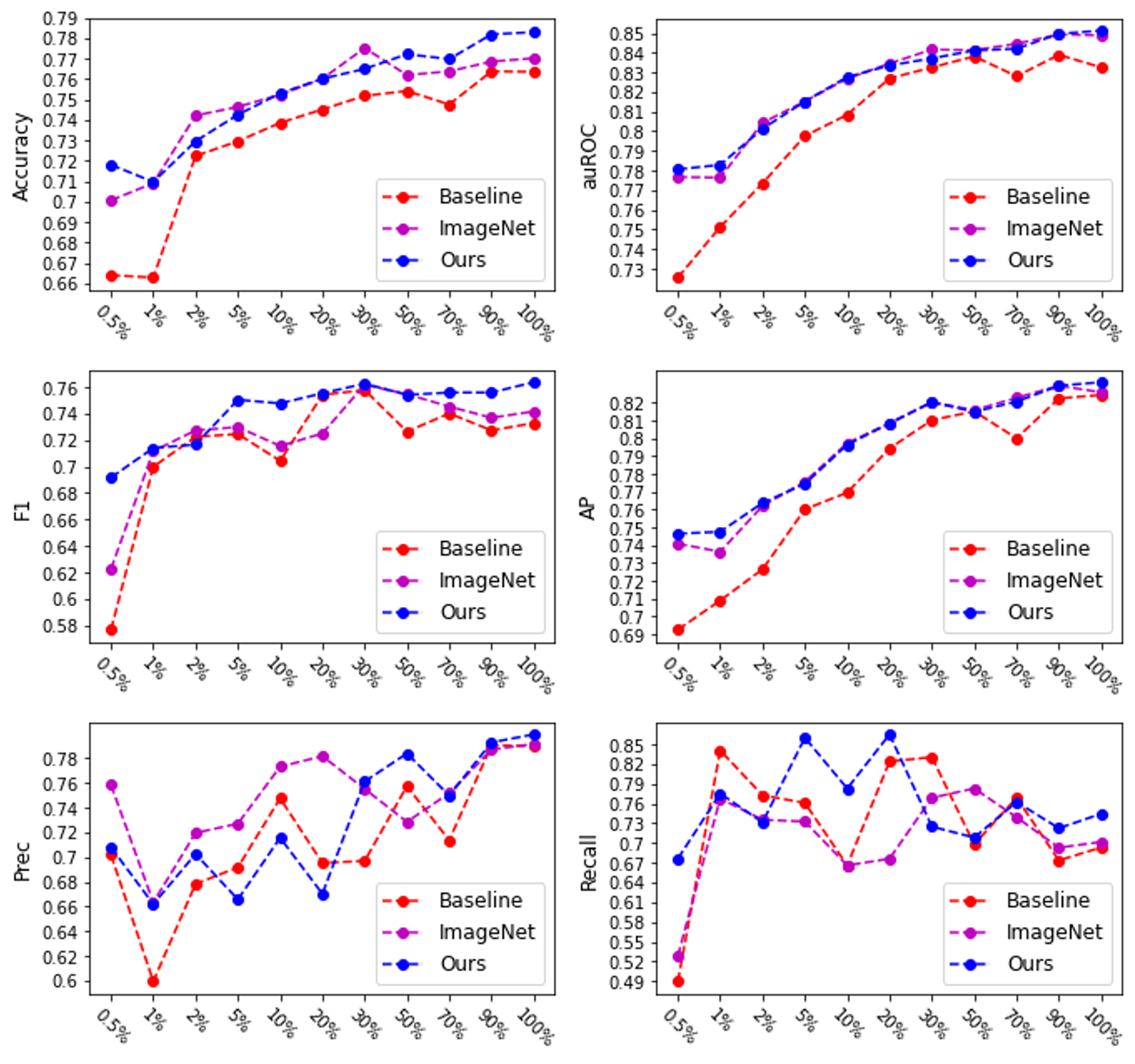}
\caption{Binary classification results on MIMIC-CXR dataset.}
\label{fig:result}
\end{figure}

Figure~\ref{fig:result} shows the result of the binary classification task on the MIMIC-CXR dataset. The results reveal that the proposed method has \ed{the best performance in general}, with better gains when only a few labeled images are available. For instance, when using 0.5\% of labeled data ($\approx$~1,800 instances), the \textit{Base} model has an accuracy of 66.41\%, \ed{\textit{ImageNet} has a 70.05\% accuracy,} while \textit{Ours} has a 71.81\% accuracy. The highest accuracy of \textit{Base} is 76.38\%, when it is trained with 90\% of the labeled data ($\approx$~325,000 instances). \ed{Both \textit{Ours} and \textit{ImageNet}} surpass the best performance of \textit{Base} with only 30\% of the training data ($\approx$~108,000 instances). Thus the need for manual labels is reduced by 67\% when using the proposed method \ed{or ImageNet pre-trained method. However, the highest accuracy of \textit{ImageNet} is 77.52\%. \textit{Ours} is able to further improve the result to 78.30\%.}

\ed{Similar trends are seen in auROC, F1, and AP scores, where the pre-trained models have better performance than \textit{Base} across all settings; the performance of the proposed method is better than the ImageNet pre-trained models in almost all of the settings. One thing worth noting is the proposed method has a consistent performance across the four metrics (accuracy, auROC, F1, and AP). However, \textit{ImageNet} has a relatively poor performance on the F1 score. For instance, when only using 0.5\% of training data ($\approx$~1,800 instances), the \textit{Base} model has an F1 score of 0.5768. \textit{ImageNet} has an F1 score of 0.6227, which is relatively low when compared with the \textit{ImageNet} performance with 0.5\% training on the other three metrics. However, \textit{Ours} has an F1 score of 0.6917 that is within the performance range of the other three metrics when using   0.5\% of training data. With 10\% of training data ($\approx$~36,000 instances), \textit{Ours} improves F1 score to 0.7477, but \textit{ImageNet}'s score is 0.7155. The F1 score of \textit{ImageNet} does reach   the same range as \textit{Ours} ($\approx 0.76$) with 30\% of training data. However, \textit{ImageNet} performance starts declining when increasing the size of the training dataset, but \textit{Ours} maintains the  same level after that.}

\ed{There is no clear winner for precision and recall, but the proposed method still scores better across most settings. For both of the metrics, \textit{Ours} wins 11 out of 22 training fractions, \textit{ImageNet} wins 6 out of 22 training fractions, and \textit{Base} wins 5 out of 22 training fractions. In addition, the figure reveals that the worst performances are generated by the \textit{Base} model with 0.5\% or 1\% of training data. The best performances are achieved by \textit{Ours}. }

\ed{One particular observation is that for very small amounts of training data, such as 0.5\% to 1\% of the full dataset ($\approx$~1,800 to 3,600 instances), TIMNet performs
consistently well across all evaluation metrics. ImageNet pre-trained model performs well on most of the metrics but scores poorly on F1 and Recall. 
The \textit{Base} model has the worst performance across all the metrics when the training data is limited.
}

\subsubsection{Multi-Label Classification}

\begin{figure}[!tb]
\centering
\includegraphics[width=0.5\textwidth]{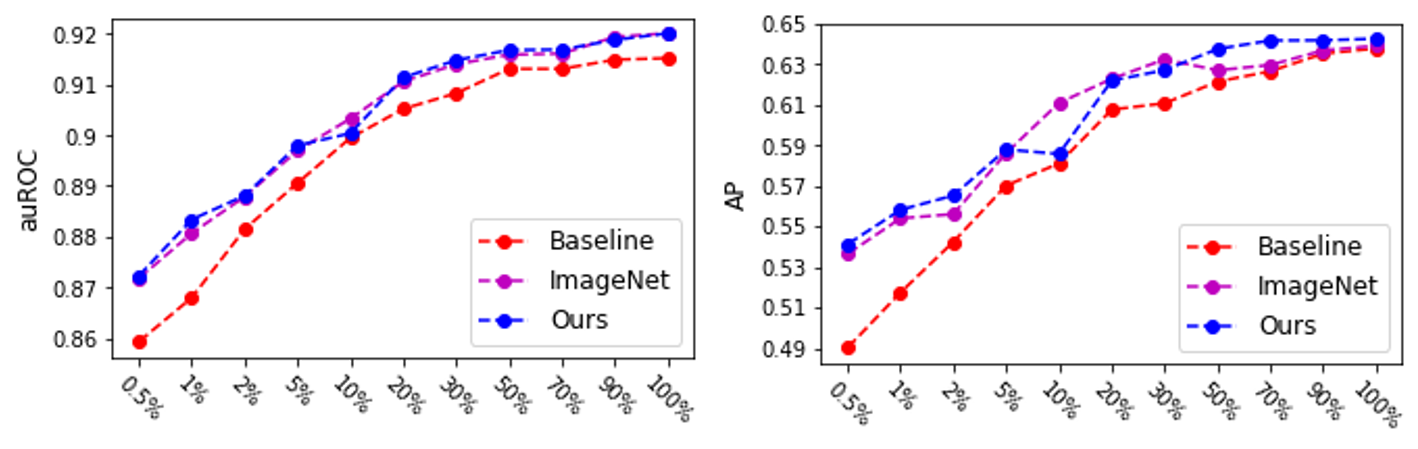}
\caption{Multi-label learning results on MIMIC-CXR dataset.}
\label{fig:result_mlmc}
\end{figure}

Figure~\ref{fig:result_mlmc}, the multi-label classification results on MIMIC-CXR, also shows the superior performance of \ed{the pre-training models} compared to the \textit{Base} model. \ed{Both} TIMNet \ed{and ImageNet pre-training are} able to significantly reduce the need for manual labels while achieving a comparable performance. The \textit{Base} model reaches its best performance (0.9152 auROC) with 100\% of the training data, while \textit{Ours} \ed{and \textit{ImageNet}} can achieve a similar performance (0.9148 \ed{and 0.9137} auROCs, \ed{respectively}) with only 30\% of the training data. \ed{With 50\% of the training data, \textit{Ours} reaches 0.9169 auROC, while \textit{ImageNet} reaches 0.9159 auROC.} 

\ed{It is hard to pick a winner when comparing \textit{Ours} against \textit{ImageNet}. Both of the methods have similar performance on this multi-label classification task, especially for auROC. Though \textit{Ours} scores better in 9 out of 11 training fractions, most of them are only marginally better than \textit{ImageNet}. Slightly larger gains are observed in AP, which may show some advantage of TIMNet pre-training.}

\subsection{Classification on an External Dataset}

This section aims to evaluate the TIMNet pre-trained weights on the dataset that is not used for the pre-training. We choose the Mendeley-V2 dataset for this evaluation because it is a pediatric pneumonia diagnosis dataset with the same medical imaging modality as MIMIC-CXR and different patient demographics.

Four models are trained and compared for pneumonia diagnosis using Mendeley-V2, namely, \textit{Ours}, \textit{Base}, \textit{ImageNet}, and \textit{C2L}~\cite{zhou2020C2L}. 
The \textit{Ours}, \textit{Base}, \ed{and \textit{ImageNet}} follow the same setup with the previous experiment, which has a TIMNet pre-trained feature extractor, a randomly initialized feature extractor, \ed{or an ImageNet pre-trained feature extractor}, respectively. The feature extractor in \textit{C2L} is pre-trained using the \textit{comparing to learn} method, the recent top method for self-supervised pre-training in medical imaging analysis tasks~\cite{zhou2020C2L}. We use the author-released PyTorch code and pre-trained weights for ResNet-18 in this study\footnote{\href{https://github.com/funnyzhou/C2L_MICCAI2020}{https://github.com/funnyzhou/C2L\_MICCAI2020}}. These weights are also trained on the MIMIC-CXR dataset. 

\begin{figure}[!tb]
\centering
\includegraphics[width=0.5\textwidth]{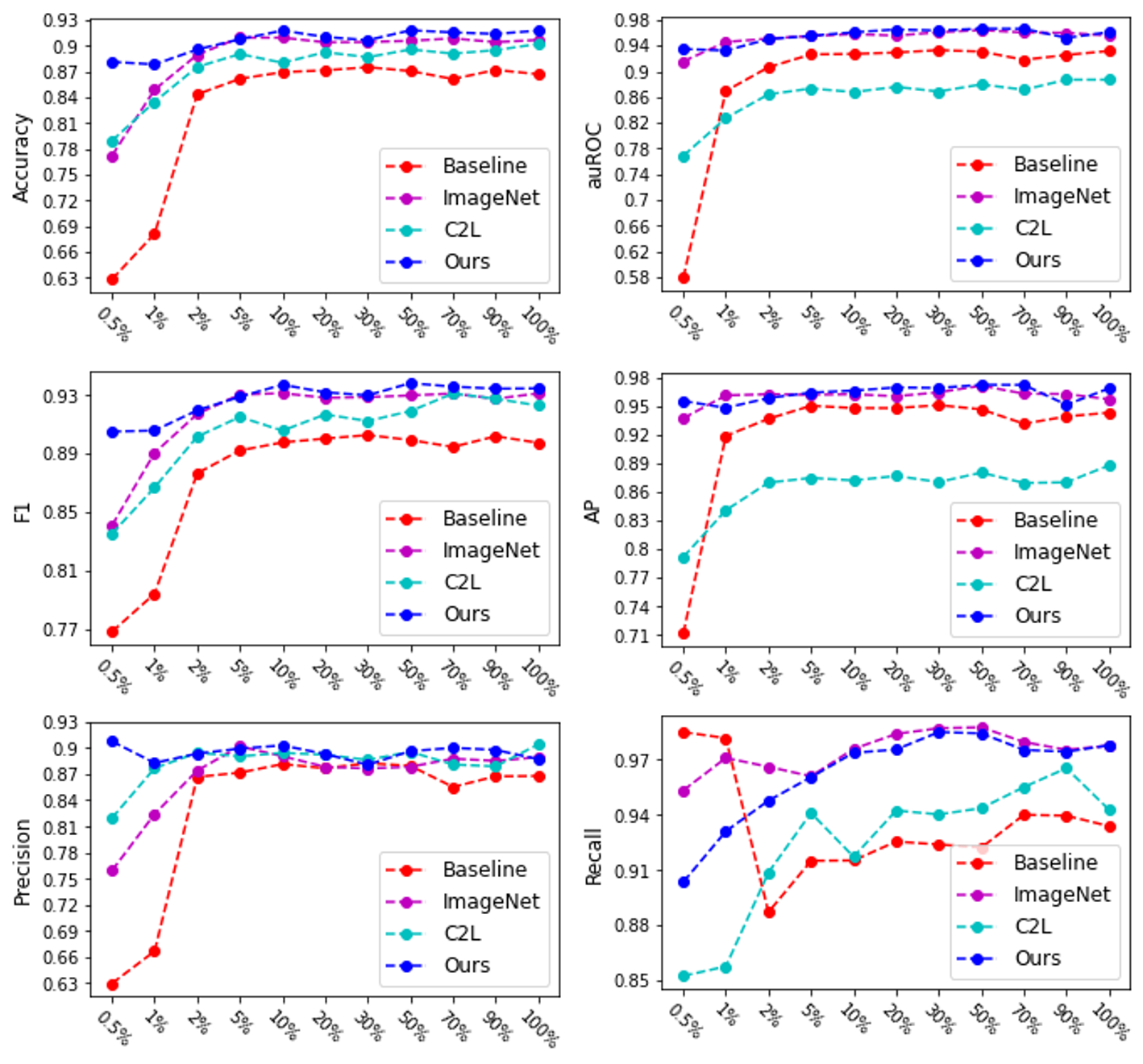}
\caption{Binary classification results on Mendeley-V2 dataset.}
\label{fig:result_mendeley}
\end{figure}

Figure~\ref{fig:result_mendeley} shows the performance of the compared models. The results reveal that TIMNet's pre-trained features work surprisingly well and achieve the best performance, in general, on this dataset among the four methods on all evaluation metrics, except Recall. It is able to reduce the need for labeled data by 98.33\% compared with the Base. The \textit{Base} model achieves its highest accuracy of 87.52\% using 30\% of the training data, and the highest auROC of 0.9333 also uses 30\% of the training data, while \textit{Ours} outperforms \textit{Base} with using only 0.5\% of training data with an 88.14\% accuracy and a 0.9352 auROC. Thus the reduction in training instances is $(30-0.5)/30=98.33\%$. The highest performances of \textit{Ours} are 91.87\% accuracy and 0.9613 auROC. These are also nearly 5\% (ACC) and 3\% (auROC) higher than those of the \textit{Base} model. 

ImageNet achieves the second overall best performance, which is close but worse than the proposed method. One particular advantage of the proposed method is when the training set is small. For instance, when using only 0.5\% of the training data, \textit{Ours} gets an approximately 88\% accuracy and an F1 score of 0.9. \textit{ImageNet} only gets about 77\% and 0.84, respectively. The difference between \textit{Ours} and \textit{ImageNet} reduces when increasing the amount of training data. On average, \textit{Ours} performed about 1 to 2\% better than \textit{ImageNet} on all evaluation metrics, except recall, on which \textit{Ours} is about 1\% worse than \textit{ImageNet}.

\textit{C2L} performs significantly better than the \textit{Base} when the training data size is extremely small. For instance, the method has about 21\% higher accuracy and 33\% higher auROC than the \textit{Base} when trained using only~0.5\% of the training data. However, a mixed result shows up when increasing the size of training data. On accuracy, F1 score,  precision, and recall, \textit{C2L} generates the third-best result among the four methods. It surpasses the best performance of \textit{Base} on accuracy using only 2\% of the training data and gets a comparable performance on F1 score using also only 2\% of the training data, resulting in a 93.33\% data reduction. However, \textit{C2L} performs worse than \textit{Base} on auROC and AP. C2L's highest auROC and AP are 0.8873 and 0.8877, respectively, which are 4.93\% or 6.66\% less than the \textit{Base} scores, respectively.

The high accuracy but low auROC and AP may imply the \textit{C2L} model favors  one class when making the prediction. The high F1 score may indicate the method works well when using 0.5 as the threshold for the binary classification task, but low auROC and AP may reveal that when changing the threshold to other values, the model performance may decrease. One feasible explanation for such a phenomenon is that as a self-supervise learning approach, \textit{C2L} may heavily rely on the assumption that features from the same image should be similar. Without providing any supervision from domain experts, the feature learned under such assumption is open-ended, which may not robust enough in some cases.

\begin{figure}[!tb]
     \centering
     \begin{subfigure}[b]{0.465\textwidth}
         \centering
         \includegraphics[width=\textwidth]{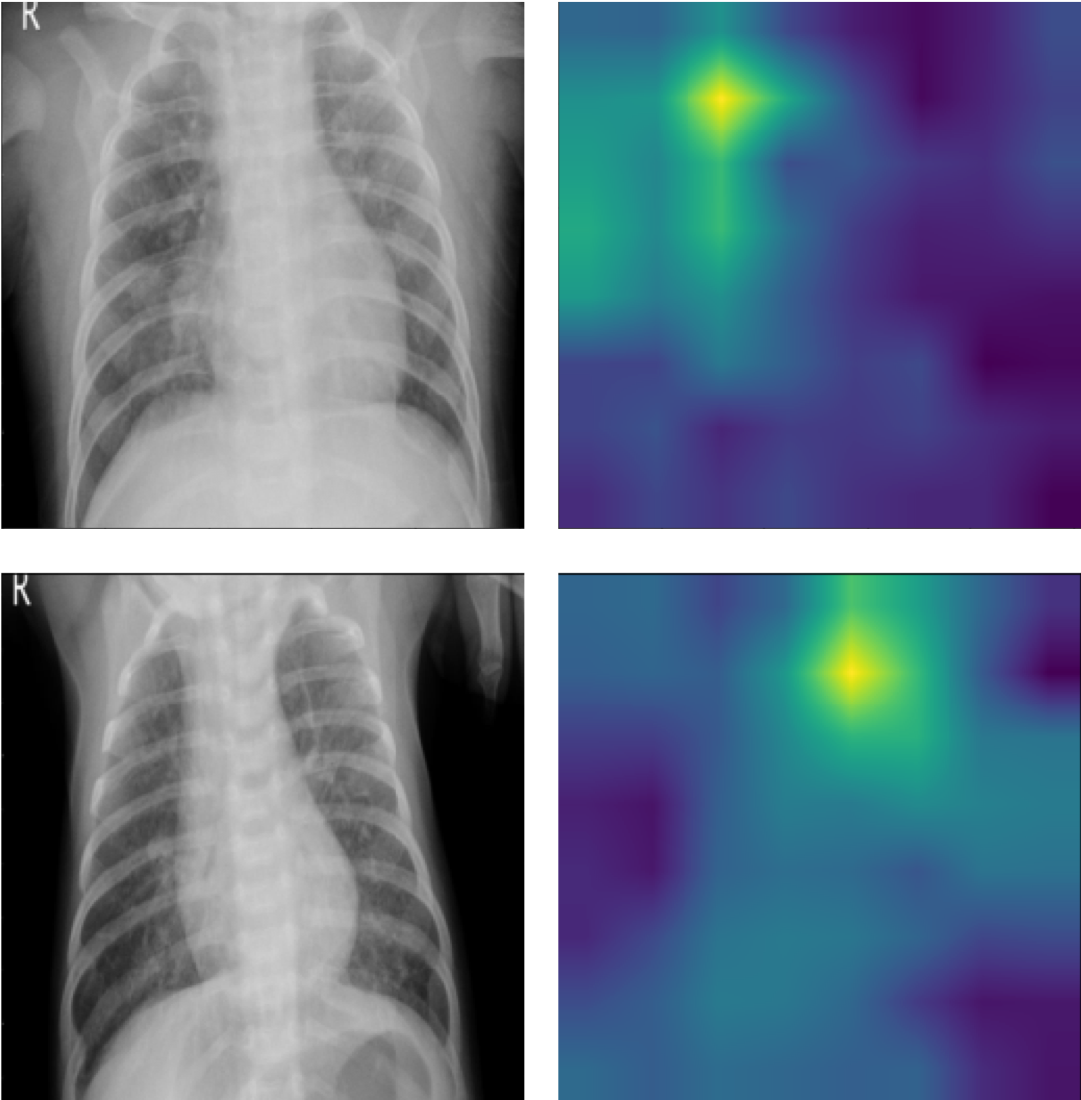}
     \end{subfigure}
    \caption{Two CAM visualizations for pediatric chest X-ray pneumonia classification on the Mendeley-V2 dataset with chest X-ray on the left and CAM on the right. The CAMs reveal that the model correctly focuses on the corresponding areas show some concerns about pneumonia.
    }
    \label{fig:cam_mendeley}
\end{figure}

Figure~\ref{fig:cam_mendeley} shows two class activation mapping (CAM)~\cite{zhou2016learning} visualizations for the pediatric pneumonia diagnosis that performed by \textit{Ours}. The pixel values in the CAMs are associated with the contribution to the classification decision. A higher value (brighter color) indicates a higher contribution to the class decision. Both cases in the figure are ground-truth positive cases. The CAMs reveal that the model correctly focuses on the corresponding areas that show some concerns about pneumonia.

\subsection{Discussion}

We proposed learning meaningful feature representations of medical images using TIMNet via a text-image matching task without acquiring additional manually labeled annotations. Our method can help build effective models for downstream predictions that rely only on image input. During our experiments, we discovered that better text-image matching performances usually lead to improved downstream application performances in terms of higher accuracy and reduced need for labeled images. 

\begin{figure}[!tb]
     \centering
     \begin{subfigure}[b]{0.495\textwidth}
         \centering
         \includegraphics[width=\textwidth]{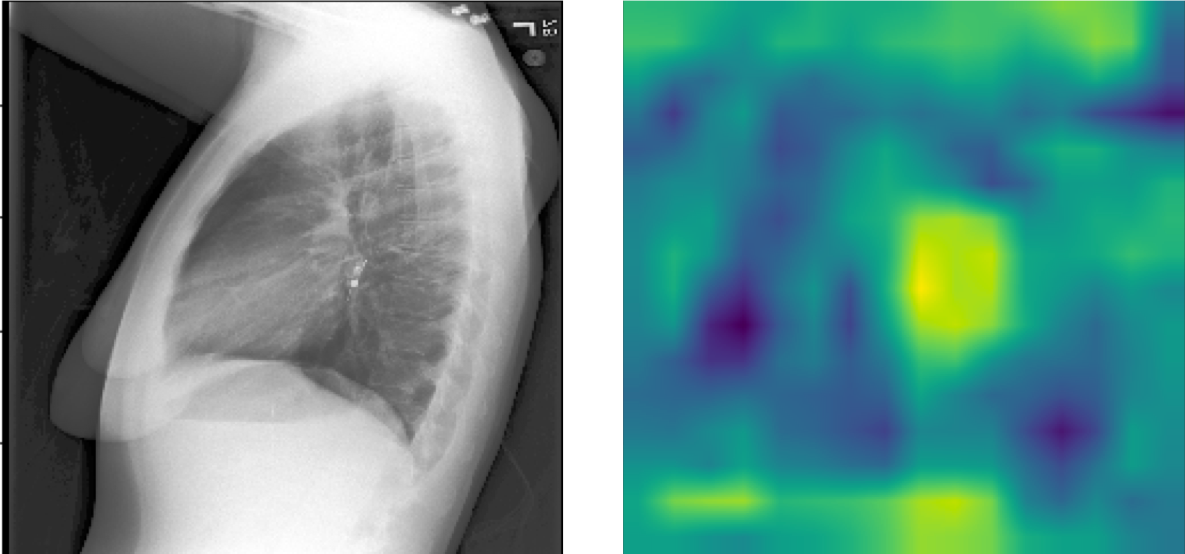}
         \caption{FINDINGS: The lungs are clear. There is no pneumothorax nor effusion. Cardiomediastinal silhouette is within normal limits. Radiopaque densities seen in the mid to distal esophagus with additional focus just past the GE junction. This may represent patient's esophageal pH probe. \newline}
         \label{fig:tim_cam_1}
     \end{subfigure}
     \hfill
     \begin{subfigure}[b]{0.465\textwidth}
         \centering
         \includegraphics[width=\textwidth]{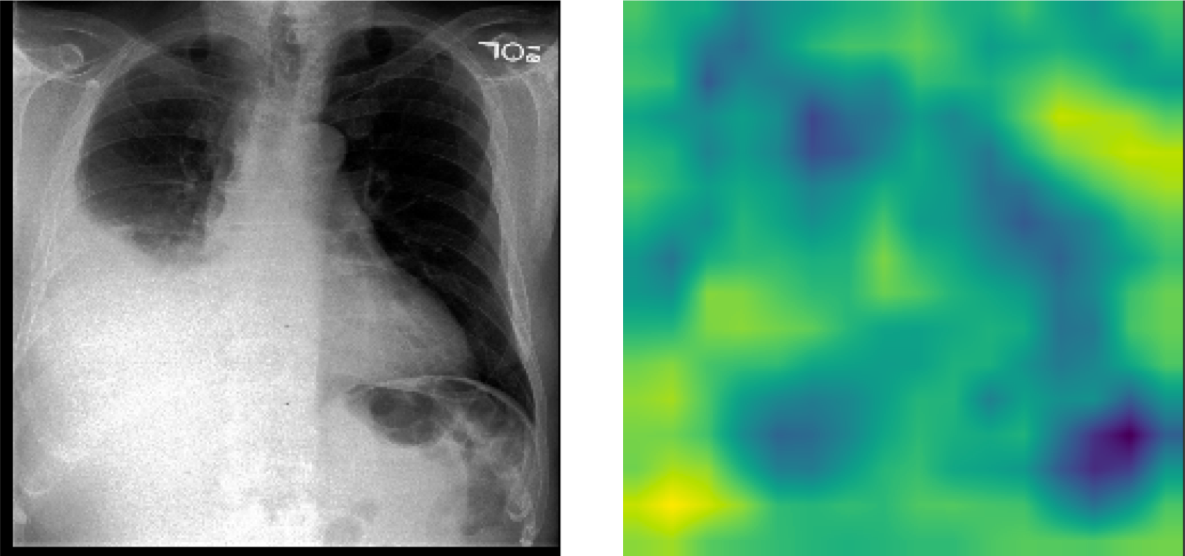}
         \caption{FINDINGS: The cardiac, mediastinal and hilar contours appear unchanged. There is no shift of mediastinal structures. There is a large right-sided pleural effusion, which has increased since the earlier radiographs and perhaps slightly since the more recent CT. There is no pneumothorax. The left lung remains clear. \newline}
         \label{fig:tim_cam_2}
     \end{subfigure}
     \begin{subfigure}[b]{0.465\textwidth}
         \centering
         \includegraphics[width=\textwidth]{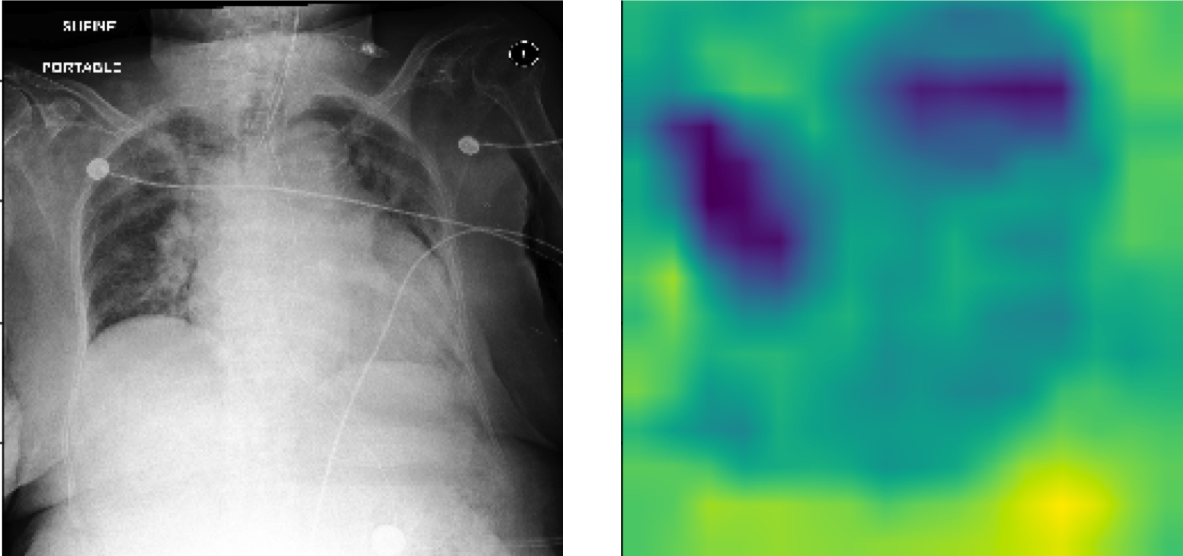}
         \caption{FINDINGS: ET tube is seen with tip approximately 1.8 cm from the carina. Enteric tube seen passing below the inferior field of view. Lower lung volumes are noted on the current exam with bilateral parenchymal opacities which could be due to edema or infection. Prominence of the right hilum is again noted. Moderate cardiomegaly and appears to have progressed since prior could potentially being part due to changes in positioning. No acute osseous abnormalities. Surgical clips project over the left chest wall/axilla.}
         \label{fig:tim_cam_4}
    \end{subfigure}
    \caption{CAM visualizations of text and image matching on MIMIC-CXR with chest X-ray on the left and CAM on the right. The CAMs reveal that the model focuses on the corresponding areas that show some concerns in the textual findings.}
    \label{fig:cam_result}
\end{figure}

The pre-trained TIMNet used in this study has a text-image matching performance of 74\% accuracy and 0.83 auROC. Table~\ref{table:text_image_matching} shows the relevant results of the evaluation. 
Figure~\ref{fig:cam_result} shows CAM visualizations of TIMNet on the text-image matching task. The findings of the radiology reports are displayed below the images. The CAMs suggest that the decisions made by TIMNet are reasonable. For instance, in 
Figure~\ref{fig:tim_cam_1}, the radiology report mentions radiopaque densities in the mid to distal esophagus, and the CAM appears to show that in the middle part of the image. For
Figure~\ref{fig:tim_cam_2}, the radiology report indicates increased right-sided pleural effusion, and CAM shows more significant contributions near the effusion areas on the right-hand side of the figure. Figure~\ref{fig:tim_cam_4} shows a similar correspondence between the CAM-based image segment contributions to the model decision and the textual report.

\begin{table}[!tb]
	\centering
	\addtolength{\tabcolsep}{-2.5pt}
	\caption{Text and image matching results}
	\begin{tabular}{|c|c|c|c|c|c|c|c|}
    \hline
    \textbf{Dataset} & \textbf{Accuracy} & \textbf{auROC}  & \textbf{F1 Score}  & \textbf{Prec}  & \textbf{Recall}  & \textbf{AP} \\ \hline
    \textbf{MIMIC-CXR} & $0.74$ & $0.83$ & $0.74$ & $0.67$ & $0.82$ & $0.77$ \\
    \hline
  \end{tabular}
  \label{table:text_image_matching}
\end{table}

We also observed that text-image matching tasks in the medical domain are typically \textbf{more challenging} than in the natural imaging domain, specifically due to the following situations: 
\begin{itemize}
    \item Textual data in the medical domain usually contain  \textbf{more words} per instance than the natural imaging domain. The vast majority of natural image captions contain 7 to 15 words~\cite{hu2020vivo}. However, this can be ten times or higher in the medical imaging domain~\cite{johnson2019mimic}.
    
    \item Textual descriptions in the medical domain usually convey \textbf{more uncertainty and ambiguity} than those in the natural imaging domain. For example, the findings of Figure~\ref{fig:tim_cam_4} are, \textit{``moderate cardiomegaly and appears to have progressed since prior \textbf{could potentially} being part due to changes in positioning."} Such hedge statements are usually not found in natural imaging captions.
    
    \item Textual data in the medical domain usually includes \textbf{information not explicitly present} in the image, such as the findings of Figure~\ref{fig:tim_cam_2}, which mention that \textit{``the cardiac, mediastinal and hilar \textbf{contours appear unchanged}.}" Without comparison with previous studies, the network may not be able to link the words ``\textit{\textbf{contours appear unchanged}}" to any regions in the image.
    
    \item Textual data in the medical domain usually contain  \textbf{redundant and uninformative text}. The majority of words of a report could be describing situations that convey normal health, which may not be very helpful when making the decision in text-image matching. Similar normal health condition statements may appear in many examples.
\end{itemize}

These nuances make text-image matching a difficult task in the medical imaging domain. 

\section{CONCLUDING REMARKS}
The main objective of our work is to demonstrate the potential of the clinician-authored textual reports that accompany most medical images in improving image-related supervised ML applications by pretraining a feature extractor without requiring additional manually labeled annotations. Such textual narratives are readily available and routinely curated as part of healthcare operations and are therefore a natural resource to leverage. The central premise of our effort is the insight that in a latent neural dense vector representation space, it may be possible to transfer linguistic signals that characterize expert summaries of images to downstream image-based tasks through weak supervision. Based on the experiments and evaluations in this paper, we believe we have successfully verified this insight for classification tasks.

At the core of our methodology is a two-branch architecture, TIMNet, that identifies whether a textual finding corresponds to the supplied image. The main purpose of text-image matching tasks is to use the textual finding as weak supervision to learn the image feature representations. In this way, the feature extractor is trained on a large and relevant dataset without requiring additional manually labeled annotations. Subsequently, the image branch is further fine-tuned for downstream supervised tasks using a small labeled dataset. Our experiments show that with the proposed method, small fractions (2\%--30\%) of the available full training data are needed to achieve the same performance as baseline models that do not exploit textual reports. Additionally, the benefits persist across datasets, which is an excellent benefit when transferring signals from models learned on deidentified textual reports (e.g., MIMIC-CXR) to other classification settings that have fewer or no textual annotations (due to HIPAA and other privacy restrictions). 

During our experiments, we discovered that better text-image matching performances usually lead to improved downstream application performances. Thus, one of our future directions will be to further innovate the matching framework to improve the associated performance. Another important future direction is to see whether our text-image matching setup can actually transfer the image signal to downstream tasks in the NLP domain for clinical text. We believe this bidirectional feedback may help in extracting named entities (e.g., drugs, comorbidities, and anatomical sites) and relations connecting such entities (e.g., adverse drug reactions) from a variety of notes. These information extraction tasks are also usually plagued by a lack of large training datasets (esp. public ones) due to strict regulatory constraints governing textual data in medicine. Overall, we hope our work spurs further interest in exploring the synergy between text and images.

\bibliographystyle{IEEEtran}  
\bibliography{bibfile}

\end{document}